\documentclass[sigconf,screen,nonacm]{acmart}
\usepackage[utf8]{inputenc} 
\usepackage[T1]{fontenc}    
\usepackage{hyperref}       
\usepackage{url}            
\usepackage{booktabs}       
\usepackage{amsfonts}       
\usepackage{nicefrac}       
\usepackage{microtype}      
\usepackage[svgnames]{xcolor}         
\usepackage{multirow}
\usepackage{longtable}
\usepackage{amsmath}
\usepackage{bbm}
\usepackage{graphicx}
\usepackage{subcaption}
\usepackage{dblfloatfix} 
\usepackage{caption}

\begin{document}


\title{IDOBE: Infectious Disease Outbreak forecasting Benchmark Ecosystem}

\author{Aniruddha Adiga$^*$}
\author{Jingyuan Chou}
\author{Anshul Chiranth}
\author{Bryan Lewis}
\affiliation{
\institution{Biocomplexity Institute, University of Virginia}
\city{Charlottesville}
\state{Virginia}
\country{USA}
}
\email{*aa5dw@virginia.edu}

\author{Ana I. Bento}
\affiliation{Department of Public and Ecosystem Health
\institution{Cornell University College of Veterinary Medicine}
\city{Ithaca}
\state{New York}
\country{USA}
}

\author{Shaun Truelove}
\affiliation{
\institution{Johns Hopkins University}
\city{Baltimore}
\state{Maryland}
\country{USA}
}

\author{Geoffrey Fox}
\author{Madhav Marathe}
\affiliation{
\institution{Biocomplexity Institute and Department of Computer Science, University of Virginia}
\city{Charlottesville}
\state{Virginia}
\country{USA}
}

\author{Harry Hochheiser}
\affiliation{
\institution{University of Pittsburgh}
\city{Pittsburgh}
\state{Pennsylvania}
\country{USA}
}

\author{Srini Venkatramanan}
\affiliation{
\institution{Biocomplexity Institute, University of Virginia}
\city{Charlottesville}
\state{Virginia}
\country{USA}
}
\email{srini@virginia.edu}

\renewcommand{\shortauthors}{Adiga et al.}

\begin{abstract}
 Epidemic forecasting has become an integral part of real-time infectious disease outbreak response. While collaborative ensembles composed of statistical and machine learning models have become the norm for real-time forecasting, standardized benchmark datasets for evaluating such methods are lacking. Further, there is limited understanding on performance of these methods for novel outbreaks with limited historical data. In this paper, we propose IDOBE, a curated collection of epidemiological time series focused on outbreak forecasting. IDOBE compiles from multiple data repositories spanning over a century of surveillance and across U.S. states and global locations. We perform  derivative-based segmentation to generate over 10,000 outbreaks covering multiple outcomes such as cases and hospitalizations for 13 diseases. We consider a variety of information-theoretic and distributional measures to quantify the epidemiological diversity of the dataset. Finally, we perform multi-horizon short-term forecasting (1- to 4-week-ahead) through the progression of the outbreak using 11 baseline models and report on their performance. In addition to standard metrics such as NMSE and MAPE for point forecasts, we include probabilistic scoring rules such as Normalized Weighted Interval Score (NWIS) to quantify the performance. We find that MLP-based methods have the most robust performance, with statistical methods having a slight edge during the pre-peak phase. IDOBE dataset along with baselines are released publicly on \url{https://github.com/NSSAC/IDOBE} to enable standardized, reproducible benchmarking of outbreak forecasting methods.


\end{abstract}



\keywords{Forecasting, Benchmark, Epidemics, Timeseries, Machine Learning}


\maketitle
\section{Introduction}
In recent years, epidemic forecasting has emerged as an active subdomain of computational epidemiology. Short-term forecasts of infectious disease activity have been adopted by various sub-national, national, and international agencies to guide outbreak response~\cite{lutz2019applying}. Agencies such as US Centers for Disease Control and Prevention (CDC) have established dedicated centers focused on improving the science, engineering, and translation of forecasting and outbreak analytics. Multi-model ensembles have been constituted to support both seasonal (e.g., Influenza)~\cite{mathis2024evaluation} and pandemic (e.g., COVID-19)~\cite{cramer2022evaluation} prediction efforts. Such efforts have been expanded to producing scenario-based projections to guide public health policy~\cite{borchering2023public,loo2024us}. Unlike projection models which mostly rely on mechanistic representations of underlying dynamics, forecast ensembles are constituted by a diverse collection of models~\cite{reich2019collaborative, adiga2021all} including machine learning, statistical and mechanistic approaches, and take advantage of multiple data streams~\cite{adiga2022enhancing} including syndromic, clinical, and environmental surveillance~\cite{lopez2023predictive}, as well as internet-based indicators. Through this partnership, robust infrastructure has been developed to undertake such Hub-style efforts for future outbreaks~\cite{shandross2025multi, kerr2025coordinating, bosse2022evaluating}. 

While significant strides have been made in advancing real-time epidemic forecasting, there is a lack of standardized, multi-disease benchmark datasets for performance evaluation of existing models and ensembles. This is especially challenging in the context of operationalizing such models for a novel outbreak~\footnote{For our purposes, we will adopt CDC's definition of outbreak as a period with more disease cases than expected for a given time, within a specific location, and for a target population~\cite{cdc_outbreak_defn}.} either in a region with limited data availability (e.g., 2014-16 West African Ebola outbreak) or limited historical or seasonal data (e.g., COVID-19 in early 2020). While multiple real-time~\cite{biggerstaff2016results, cramer2021united} and retrospective~\cite{reich2019collaborative} efforts have been undertaken, there is also need for standardized evaluation of methods outside the operational context. Most real-time efforts (outside COVID-19) have involved seasonal epidemics like Influenza, Dengue, and hence have leveraged the existence of historical data. Pre-trained epidemic models~\cite{kamarthi2023pems} trained across various regional outbreaks will be needed for rapid deployment and wider adoption.

\subsection{Contributions}
In this paper, we present IDOBE, an ecosystem for benchmarking models in the task of infectious disease outbreak forecasting. IDOBE comprises curated and preprocessed outbreaks drawn from diverse data repositories, along with a collection of baseline models and standardized evaluation metrics relevant for epidemic forecasting. Specifically:

\begin{itemize}
\item We preprocess epidemic time series datasets for \textbf{13 different diseases}, across \textbf{248 unique locations}, and outcomes such as outpatient visits, confirmed cases and hospitalizations. The dataset comprising \textbf{10799 outbreaks}, is compiled from existing disease data repositories such as Tycho~\cite{van2018project}, JHU-CSSE COVID-19 data repository~\cite{dong2020interactive}, as well public health surveillance published by US CDC and the National Healthcare Safety Network (NHSN).

\item We propose a suite of \textbf{information-theoretic} and \textbf{distributional measures} to characterize the diversity of outbreak trajectories contained in IDOBE. While some of these measures such as entropy~\cite{dalziel2018urbanization} and permutation entropy~\cite{scarpino2019predictability} have been used in isolated studies, such a multi-dimensional characterization has not been performed before in the context of epidemic outbreak trajectories.

\item We generate for multi-horizon short-term forecasting (1- to 4-week ahead) from \textbf{11 baseline models}  across the progression of the outbreak. The baseline models span a variety of statistical (ARIMA, ETS), MLP-based (MLP, N-BEATS, N-HiTS), transformer-based (Informer,TFT), and RNN-based (RNN, GRU, LSTM, TCN) methods. In addition to producing point forecasts, the models are run with uncertainty quantification to produce probabilistic forecasts in the Hubverse~\cite{kerr2025coordinating} standard format consistent with existing forecasting Hubs. 

\item In addition to evaluating the point forecasts using standard metrics such as \textbf{MAPE, NMSE}, we also incorporate a \textbf{normalized} version of the \textbf{Weighted Interval Score (NWIS)}~\cite{bracher2020evaluating} for the probabilistic forecasts. We interpret the performance of baseline models across epidemiological context of the outbreak (pre-peak or post-peak) and forecast horizon as well as by disease.
\end{itemize}

\paragraph{Data and code availability} To ensure reproducibility and facilitate further exploration, we provide preprocessed datasets, trained baseline models, and scripts for extracting outbreak analytics and evaluation metrics through the public repository: \url{https://github.com/NSSAC/IDOBE}

\begin{figure*}[ht!]
    \centering
    \includegraphics[width=0.95\linewidth]{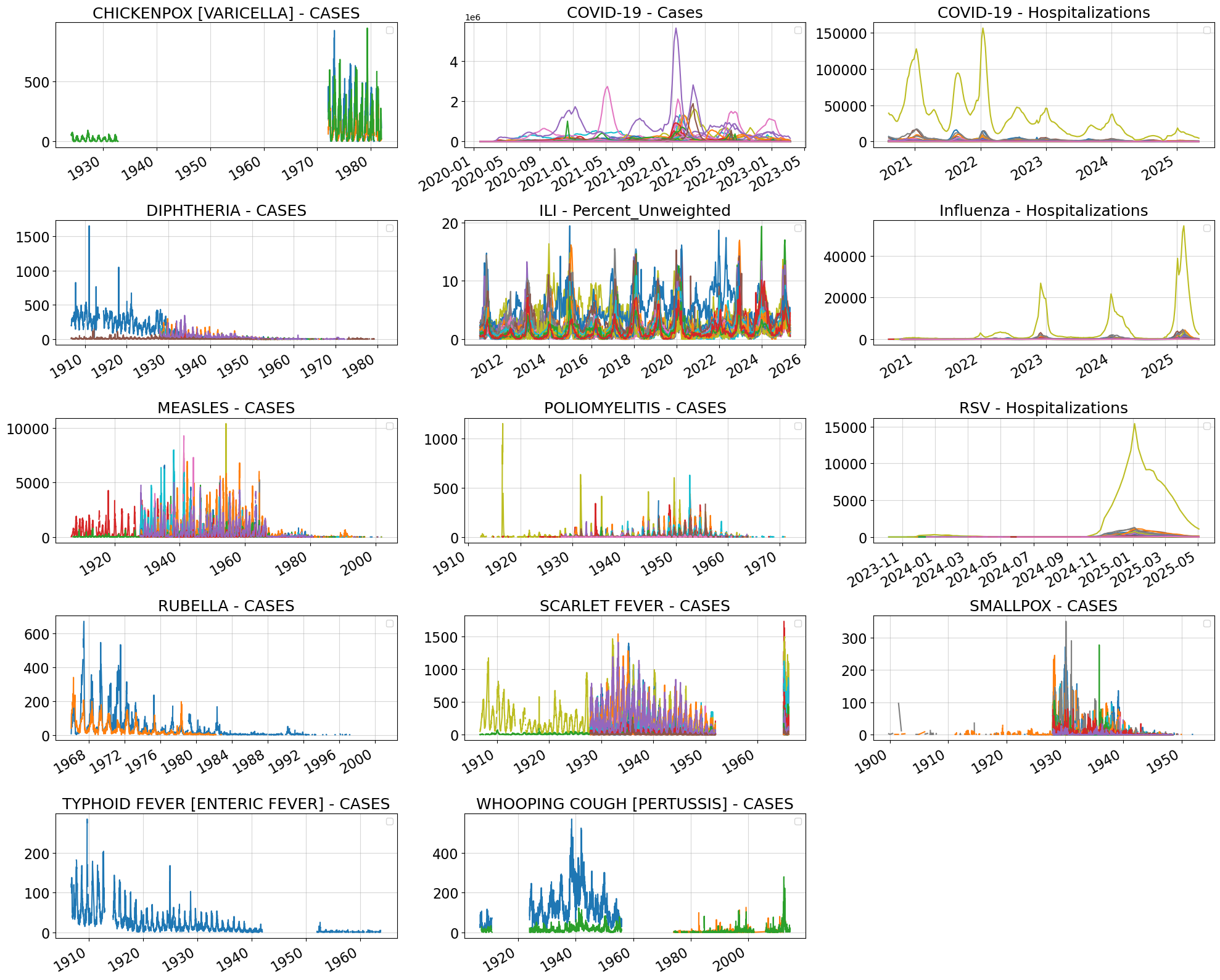}
    \caption{Timeseries corresponding to different diseases.}
    \label{fig:diseases}
\end{figure*}

\subsection{Related Work}
Collaborative forecasting ``challenges'' have been conducted for more than a decade under the Epidemic Prediction Initiative by US CDC, for targets ranging from seasonal influenza-like illness (ILI) forecasting (2013-now)~\cite{biggerstaff2016results, mathis2024evaluation}, Dengue (2015)~\cite{johansson2019open}, Chikungunya (2014)~\cite{del2018summary} and West Nile Virus~\cite{holcomb2023evaluation}. Of these, ILI forecasting has received the most attention, with retrospective forecast performance evaluation conducted as part of the FluSight Network~\cite{reich2019collaborative}. Similar efforts during the COVID-19 pandemic~\cite{cramer2021united, cramer2022evaluation} played a key role in influencing public policy, although model performance varied significantly across key epochs~\cite{rosenfeld2021epidemic, lopez2024challenges}. While most of these efforts target seasonal, recurrent epidemics, IDOBE is designed around discrete outbreak episodes, mirroring novel or emerging pathogen scenarios.

Other efforts such as the M-competitions~\cite{makridakis2000m3, makridakis2020m4, makridakis2022m5} have been undertaken in the forecasting community outside epidemiology. Recently, benchmarks such GIFT-Eval~\cite{aksu2024gift} have been developed for general time series forecasting tasks. Recently, Pre-trained~\cite{kamarthi2023pems} and Foundation models~\cite{kalahasti2025foundation} have been developed and evaluated in the epidemiological context. A similar benchmarking framework to ours was envisioned in ~\cite{srivastava2021epibench}, although it was limited to COVID-19 forecasting in the US with simpler evaluation metrics and did not tackle the broader task of outbreak forecasting as outlined in this paper.

\section{Methodology}
\subsection{Task definition}\label{sec:task}
Consider an outbreak of disease $d$ reported at location $l$ for a particular disease outcome $o$ (cases, deaths, or hospitalizations). Let $i$ be the unique identification number of the outbreak, and let $T_i$ be the duration of the outbreak. We denote the time series corresponding to the outbreak $i$ as a vector $\textbf{x}_z(0:T_i) = [x_z(0),x_z(1),\cdots, x_z(T_i)]$, where $x_z(\cdot)$ is the value of the outcome and $z=(d,l,o,i)$. In this work, we focus on simulating the task of short-term forecasting of outbreaks in real-time. Thus, given observations up to $u$, denoted as $\textbf{x}_z(0:u)$, the goal is to forecast the values $\textbf{x}_z(u+1:u+h)$, where $h$ is the forecast horizon. In epidemic forecasting, similar to weather forecasting~\cite{gneiting2007probabilistic}, predictive probability distribution of future values is the standardized format for reporting forecasts. Hence, the real-time forecasting task would involve learning a model $f_\Theta (k): \textbf{x}_z(0:u) \rightarrow P(x_z(u+k) | \textbf{x}_z(0:u), \Theta)$, for $0\leq u\leq T_i$, $0 \leq k \leq h$.


The model parameters $\Theta$ are typically learned from historically observed outbreaks. Typically, historical data consists of multiple outbreaks. In the subsequent sections, we discuss the process of extracting individual outbreaks from the complete time series.


\subsection{Benchmark data curation}

\begin{table*}[ht!]
\small
    \centering
    \caption{Statistics of source datasets and outbreaks across different diseases}
    \begin{tabular}{lllrllr}
\toprule
 disease & event & \# outbreaks & start date & end date & \# years \\
\midrule
CHICKENPOX [VARICELLA] & CASES & 25 & 1972-02-19 & 1982-08-07 & 10 \\
COVID-19 & CASES & 981 & 2020-03-21 & 2025-05-10 & 5 \\
COVID-19 & HOSPITALIZATIONS & 473 & 2020-07-18 & 2025-05-31 & 5 \\
DIPHTHERIA & CASES & 1062 & 1927-12-03 & 1978-11-18 & 51 \\
ILI & PERCENT UNWEIGHTED & 867 & 2010-09-11 & 2025-05-31 & 15 \\
INFLUENZA & HOSPITALIZATIONS & 337 & 2020-10-24 & 2025-05-31 & 5 \\
MEASLES & CASES & 1517 & 1927-12-24 & 1997-05-17 & 69 \\
POLIOMYELITIS & CASES & 2442 & 1921-12-31 & 1971-08-21 & 50 \\
RSV & HOSPITALIZATIONS & 55 & 2023-11-11 & 2025-05-31 & 2 \\
RUBELLA & CASES & 84 & 1966-10-01 & 1997-03-08 & 30 \\
SCARLET FEVER & CASES & 1333 & 1927-11-05 & 1967-02-04 & 39 \\
SMALLPOX & CASES & 1324 & 1910-08-06 & 1949-05-07 & 39 \\
TYPHOID FEVER [ENTERIC FEVER] & CASES & 48 & 1928-04-21 & 1963-06-29 & 35 \\
WHOOPING COUGH [PERTUSSIS] & CASES & 251 & 1937-03-20 & 2015-05-09 & 78 \\
\bottomrule
\end{tabular}
    \label{tab:stats}
\end{table*}

\begin{figure*}[ht!]
    \centering
    \includegraphics[width=0.9\linewidth]{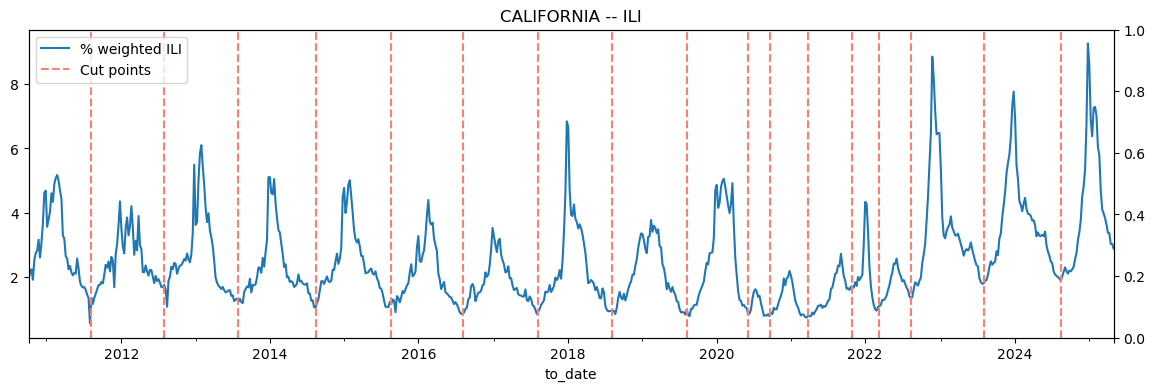}
    \caption{Example segmentation of timeseries into individual outbreaks (pink vertical dashed lines indicate the cutpoints).}
    \label{fig:segment}
\end{figure*}

\subsubsection{Available datasets}
We collect disease-specific data from four different sources: Tycho~\cite{van2018project}, JHU-CSSE~\cite{dong2020interactive}, US CDC, and NHSN. Each data source has data collected over different timelines, cover different diseases, temporal resolution (daily or weekly), and locations and the data format and nomenclature used also vary. Daily counts (only available for COVID-19 cases) are often dominated by reporting noise and day-of-the-week effects and can be smoothed out by aggregating daily data to weekly resolution. Moreover, many epidemic forecasting efforts require weekly forecasts~\cite{mathis2024evaluation, cramer2021united}. Accordingly, we align the temporal resolution across all datasets to be weekly, indexed by the MMWR week (Sunday-Saturday). Since COVID-19 confirmed cases are reported as daily counts, weekly count was obtained by summing the reported counts from Sunday-Saturday.

The Tycho dataset~\cite{van2018project} contains weekly reports of 56 infectious diseases collected between 1888 and 2014 across various U.S. cities, counties, and states. However, no single disease was reported continuously throughout the entire interval. Among the Tycho time series, many contain substantial missing data. In order balance retention of long historical series with the need for sufficient observed data for model training, we exclude timeseries with significant proportion of missing values and no prominent outbreak segments. For the remaining timeseries, missing values are imputed using linear interpolation. After filtering, nine diseases from the Tycho dataset were retained.

The Johns Hopkins University (JHU) COVID-19 data repository~\cite{dong2020interactive} provides time series of reported COVID-19 cases globally, as well as across U.S. states and counties. At the global level, we include 201 locations, comprising countries and a few special administrative regions or events. At the U.S. level, we focus on all 50 states. Although the JHU dataset includes county-level data, it is often sparse and noisy due to low case counts; thus, we exclude it in this version. We plan to incorporate county-level data in a future release. We obtained the percentage of outpatient Influenza-Like Illness (ILI) visits data from CDC at the state level for United States, spanning fifteen seasons from 2010 to 2025~\footnote{https://gis.cdc.gov/grasp/fluview/fluportaldashboard.html}. The National Healthcare Safety Network (NHSN) dataset~\footnote{https://www.cdc.gov/nhsn/psc/hospital-respiratory-dashboard.html} includes weekly records of new hospital admissions due to COVID-19, Influenza, and RSV across the 50 states of U.S. In total, we compile time series data for 13 diseases across the four data sources. The timeseries corresponding to different diseases are shown in Figure~\ref{fig:diseases}. 


\subsubsection{Outbreak Segmentation}  Disease surveillance corresponding to every disease $d$, location $l$, outcome $o$ consist of multiple outbreaks. We denote the historical reports as $X_{d,l}^{o}(t)$. In the benchmark dataset, we extract the individual outbreaks from each $X_{d,l}^{o}(t)$ using a segmentation function $S_\phi:X_{d,l}^{o}(t) \rightarrow \{x_{d,l,o}(t_n:t_n+T_n)\}_{n=0}^{N-1}$. Each individual outbreak is assigned a unique identifier $i$ and stored in a dictionary $\mathcal{X} = \{\textbf{x}_z\}_{z\in Z}$, where $z=(d,l,o,i)$ and $Z$ is the set of all tuples.

We employ a derivative-based function as $S_{\phi}$ to obtain individual outbreaks. We use a Python toolbox EpidemicKabu~\footnote{https://pypi.org/project/EpidemicKabu/}\cite{galvis2024epidemickabu}, which is designed to identify epidemic waves by detecting peaks, valleys, and inflection points in time series data. We employ the wave identification functionality to segment a given time series into different waves. Wave detection involves ($i$) smoothing the timeseries using a Gaussian kernel, ($ii$) determining cut points in the smoothed first derivative where the first derivative crosses the x-axis from negative to positive, and ($iii$) selecting the detected cut points whose second derivative is less than a threshold. The output of EpidemicKabu consists of a list of cut points, where each segment between two consecutive cut points is interpreted as a potential outbreak. An example of the resulting outbreak segmentation is shown in Figure~\ref{fig:segment}. 

Such a segmentation is central to the task of outbreak forecasting for multiple reasons. Typical multi-wave patterns in epidemic time series emerge due to a combination of seasonal, geographical, demographical, and biological contexts. For example, influenza epidemics exhibit strong seasonality and are often characterized by staggered peaking dynamics of different age groups. COVID-19 pandemic saw the emergence of multiple variants that resulted in distinct waves. Further, non-pharmaceutical interventions and heterogeneous population mixing could result in geographic diversity of epidemic spread that may manifest as distinct ``modes'' in an aggregate epidemic curve. These characteristics are often muted in the case of a novel outbreak, and hence a robust forecasting framework must be capable of leveraging latent dynamics within an isolated outbreak trajectory.

We discard segments whose duration is less than 8 weeks or greater than 52 weeks for two reasons: (i) to minimize detection of brief spikes as outbreaks, (ii) to separate out extremely long multi-seasonal trends.  Additionally, to ensure that sufficient context is available around each outbreak, we append four weeks of time series data both before the start and after the end of each identified segment, which result in overlapping segments across outbreaks. This provides sufficient context for models trained on fixed windows, and also helps avoid boundary effects in feature extraction. Metadata provided per outbreak includes region (state/country), time period, disease ontology, and indicators of sporadic/seasonal patterns. Table~\ref{tab:data_dict} provides the data dictionary provided to the user within the dataset.
\begin{table}[t!]
    \caption{Data dictionary describing the metadata and structure of the dataset.}
    \centering
    \resizebox{0.9\linewidth}{!}{%
    \begin{tabular}{|p{2cm}|p{5cm}|}
    \toprule
         \texttt{unique\_id} & An unique identification number for each outbreak \\
         \texttt{disease} & Type of disease (COVID-19, influenza, Smallpox, etc.)\\
         \texttt{location} & Name of the location (US states, countries, etc.) \\
         \texttt{event} & Type of burden indicator (cases, hospitalizations, etc.) \\
         \texttt{start\_date} & Start date of the outbreak (end-of-week Saturday date) \\
         \texttt{end\_date} & End date of the outbreak (end-of-week Saturday date) \\
         \texttt{duration} & Duration of the outbreak (in weeks) \\
         $[0-59]$ & Values observed for the particular outbreak for given week (counts, \%)\\  
         \bottomrule
    \end{tabular}
    \label{tab:data_dict}
    }
\end{table}



\subsection{Analytical measures}
Since the dataset consists of diverse set of outbreaks collected over different periods, under different reporting strategies, and geographical locations, we analyze the characteristics of outbreaks using multiple statistical measures. These measures capture the uncertainty/noise and shapes of the different outbreaks.

\paragraph{Entropy analysis} Following the definition of \emph{epidemic intensity} defined in~\cite{dalziel2018urbanization}, we compute the Shannon entropy of the incidence distribution of each outbreak, such that it is minimized when incidence is spread evenly across weeks and increases as incidence becomes more intensively focused in particular weeks. The incidence curve of an outbreak $i$ is normalized, that is, $\mathbf{\bar{x}}_z(0:T_i)=\frac{1}{\sum_{t=0}^{T_i}x_z(t)}\mathbf{x}_z(0:T_i)$ to obtain the incidence distribution. The Shannon entropy is computed over the probability distribution. 

\paragraph{Permutation Entropy analysis} We employ permutation entropy (PE)~\cite{bandt2002permutation} to characterize the diversity of short-term ordinal patterns within individual outbreaks. PE is a model-free measure of uncertainty in a signal and has been used to quantify the uncertainty and predictability of epidemic time series~\cite{scarpino2019predictability}. In contrast to the Shannon entropy, PE does not consider the frequency of state changes but the frequency of permutation patterns (ordinal patterns) within a signal and is characterized by an embedding dimension (order) and a delay parameter. We consider embedding dimension of 3, referring to the number of consecutive data points in a time series that are grouped to form the embedding vector. This, in turn, determines the number of ordinal patterns that can be obtained (for an embedding dimension $d=3$, the number of possible patterns would be $d!=3!$). The delay parameter determines the temporal resolution at which the patterns are analyzed and we fix the delay to be 1 week (also known as no skip) in our analysis. Signals with high stochasticity (pure white noise) will likely have all patterns occur with equal frequency and are characterized by a high PE value. On the other hand, a perfectly periodic signal will typically have low entropy. We refer the readers to \cite{bandt2002permutation} for more details, insights, and examples of PE and its parameters. 

\paragraph{Skewness and Kurtosis} These statistical measures obtained as the third and fourth moments of the incidence distribution help characterize the outbreak shape in terms of  asymmetry (skewness) and tailedness/peakedness (kurtosis) relative to a normal curve. 

\subsection{Baseline methods}
We evaluate four types of forecasting methods as baselines in this study: i). Statistical Methods, including ARIMA (Autoregressive Integrated Moving Average) and ETS (Exponential Smoothing)~\cite{hyndman2018forecasting, garza2022statsforecast}. ii). Recurrent neural network (RNN)-based methods, including GRU (Gated Recurrent Unit)~\cite{cho2014learning}, LSTM (Long Short Term Memory)~\cite{hochreiter1997long}, RNN (Recurrent Neural Network)~\cite{elman1990finding}, and TCN (Temporal Convolution Network)~\cite{lea2017temporal}. iii). Transformer-based methods, including Temporal Fusion Transformer (TFT)~\cite{lim2021temporal}, and Informer~\cite{zhou2021informer}. iv) Multilayer Perceptron (MLP)-based methods: including vanilla MLP~\cite{goodfellow2016deep}, Neural Basis Expansion Analysis for Time Series Forecasting (NBEATS)~\cite{oreshkin2019nbeats}, and Neural hierarchical interpolation for time series forecasting (NHITS)~\cite{challu2023nhits}. All statistical and deep-learning model implementations are based on the unified and efficient forecasting framework provided by \textit{Nixtla's} \texttt{StatsForecast}~\cite{garza2022statsforecast} and \texttt{NeuralForecast}~\cite{olivares2022library_neuralforecast} libraries. The motivation for using \textit{Nixtla} is that it offers a unified API for a wide range of state-of-the-art statistical and deep learning models and robust support for hyperparameter tuning via Optuna or Ray. Its modular design and scalability make it well-suited for benchmarking diverse forecasting methods under consistent experimental settings. 

All the models are trained to produce probabilistic forecasts. Nixtla’s ARIMA and ETS produce uncertainty through parametric predictive distributions implied by the fitted statistical models. Forecast variances are derived analytically from estimated model residuals assuming Gaussian errors and propagated across horizons to obtain prediction intervals and quantiles. Nixtla’s deep learning models generate uncertainty through quantile regression, training neural networks with multi-quantile (pinball) losses to predict multiple forecast quantiles without assuming an explicit data-generating distribution. We set the number of quantiles to 23 consistent with the recommendations from the FluSight and COVID-19 forecast hubs~\cite{mathis2024evaluation,cramer2021united}.




\subsubsection{Model Training/Fitting and Testing}
To accommodate the different classes of baselines, we adopt a model-class-specific training/fitting strategies. Since all models are intended for real-time deployment (Section~\ref{sec:task}), they are designed to operate under an expanding window setting (see timeseries cross-validation technique in~\cite{hyndman2018forecasting}). Specifically, as an outbreak progresses, models have access to an expanding set of observations $\textbf{x}_z(0:u)$, $7\le u\le T_i-h, \forall z$. We impose a minimum window of eight and a maximum of $T_i-h$ for training/fitting of the models.  

In the case of statistical models (ARIMA, ETS), for a given $z$, a model is fit on each of the expanding-window of observations $\textbf{x}_z(0:u)$, $7\le u\le T_i-h$. The testing is performed by generating out-of-sample forecasts from the fitted model for horizons $(u+1:u+h), 7\le u\le T_i-h$ and evaluating the forecasts against the observed values. We employ the \texttt{AutoARIMA} model provided by Nixtla to automatically select the optimal ARIMA parameters.


For the remaining model classes that require explicit training, we adopt a unified training/validation/testing strategy. Rather than splitting data along the temporal dimension, we partition the dataset across outbreaks. We shuffle the unique\_ids and split them into training/validation/testing sets in a 60\%/20\%/20\% fashion, respectively. Consequently, during testing, models are evaluated on completely unseen outbreaks. 

During training, we tune key hyperparamters and select the optimal model parameters based on the model's performance on the validation set. We employed the hyperparameter optimization framework \textsf{Optuna}~\cite{akiba2019optuna} to determine the optimal set of hyperparameters. Following are the set of key tunable hyperparameters along with the search space:

\begin{itemize}
  \item \texttt{input\_size}, which controls the length of the historical look-back window used as input to the model, sampled as an integer between $[8,\texttt{max\_input\_size}]$.

  \item \texttt{learning\_rate}, which specifies the optimizer step size during model training and is sampled on a logarithmic scale between $10^{-4}$ and $10^{-2}$.

  \item \texttt{batch\_size}, which specifies the number of samples processed per training optimization step and is selected from the set $\{16, 32, 64\}$.

\end{itemize}

Forecasting on the full test set requires generating $48\times 4 = 192$ forecasts for each of the ~2000 outbreaks and is time consuming. To enable efficient evaluation,  we randomly sample minibatches of 100 outbreaks from the test set and compute the forecast performance on each minibatch. This procedure is repeated multiple times, and final performance metrics are obtained by averaging the performance across minibatches. 

\subsection{Forecast Evaluation Metrics}
\subsubsection{Point Forecasts}
Although our baseline models are designed to provide probabilistic forecasts, to lower the barrier for entry for new models, we also include evaluation metrics for point forecasts. 
Specifically, we include normalized versions of error metrics to be scale-agnostic across the outbreaks. We consider mean absolute percentage error (MAPE) and normalized mean square error (NMSE)~\cite{tabataba2017framework} obtained for each forecast target, and averaged across outbreaks, timepoints, and horizons. For the baseline model forecasts, we assumed the values corresponding to the median (quantile level = 0.5) as the point forecasts.

\subsubsection{Probabilistic Forecasts}
\label{app:forecast_eval}
In order to compare the forecast quantiles of the different models, we use the Weighted Interval Score (WIS), the de facto standard in epidemiological forecasting community for probabilistic forecast evaluation \cite{bracher2020evaluating}:

\begin{eqnarray}
&WIS_{\alpha_{0:k}} (F,y) = \nonumber\\ &\frac{1}{K+0.5} \sum_{k=0}^{K} \frac{\alpha_k}{2} (u_k-l_k) + \frac{2}{\alpha_k} (l_k-y) \mathbbm{1}(y<l_k) + \nonumber\\&\frac{2}{\alpha_k} (y-u_k) \mathbbm{1}(y>u_k)
\end{eqnarray}

where $y$ is the observed value (ground truth case count corresponding to a week) for a given location and date, $F$ is the forecast defined in terms of the median $m$, upper quantiles $u_k$ and lower quantiles $l_k$ of the predictive distribution, respectively. $K$ is the number of intervals considered, which in our case $K=11$. Since WIS is scale dependent, we divide the obtained WIS for a given target by its ground truth value, hence yielding a normalized weighted interval score (NWIS).




\section{Results}
\begin{figure*}[]
\centering
\begin{subfigure}[t]{0.3\textwidth}
\centering
    \includegraphics[width=\textwidth]{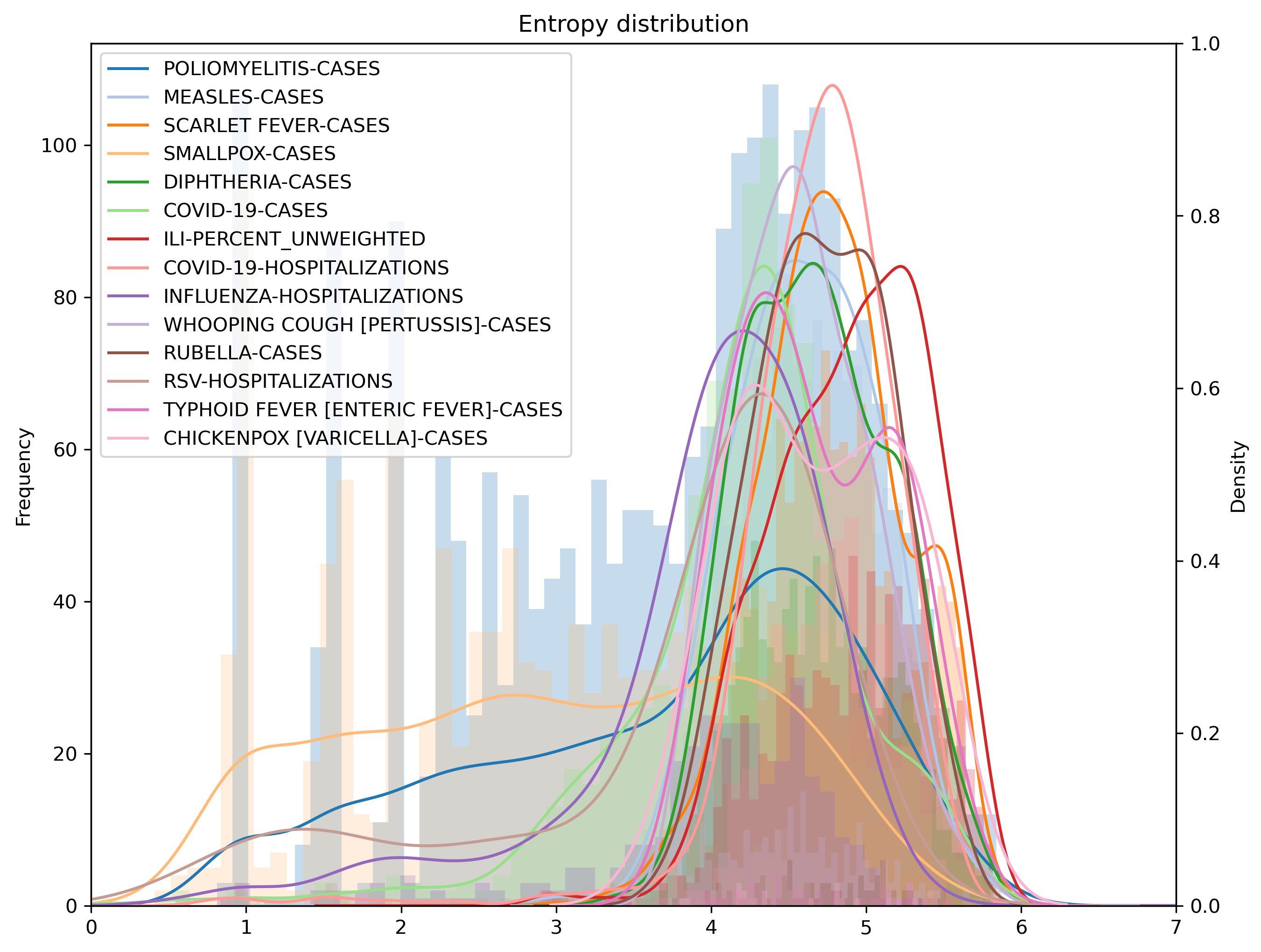}
    \caption{Entropy distribution}
    \label{fig:entropy}    
\end{subfigure}
\hfill 
\begin{subfigure}[t]{0.3\textwidth}
\centering
    \includegraphics[width=\textwidth]{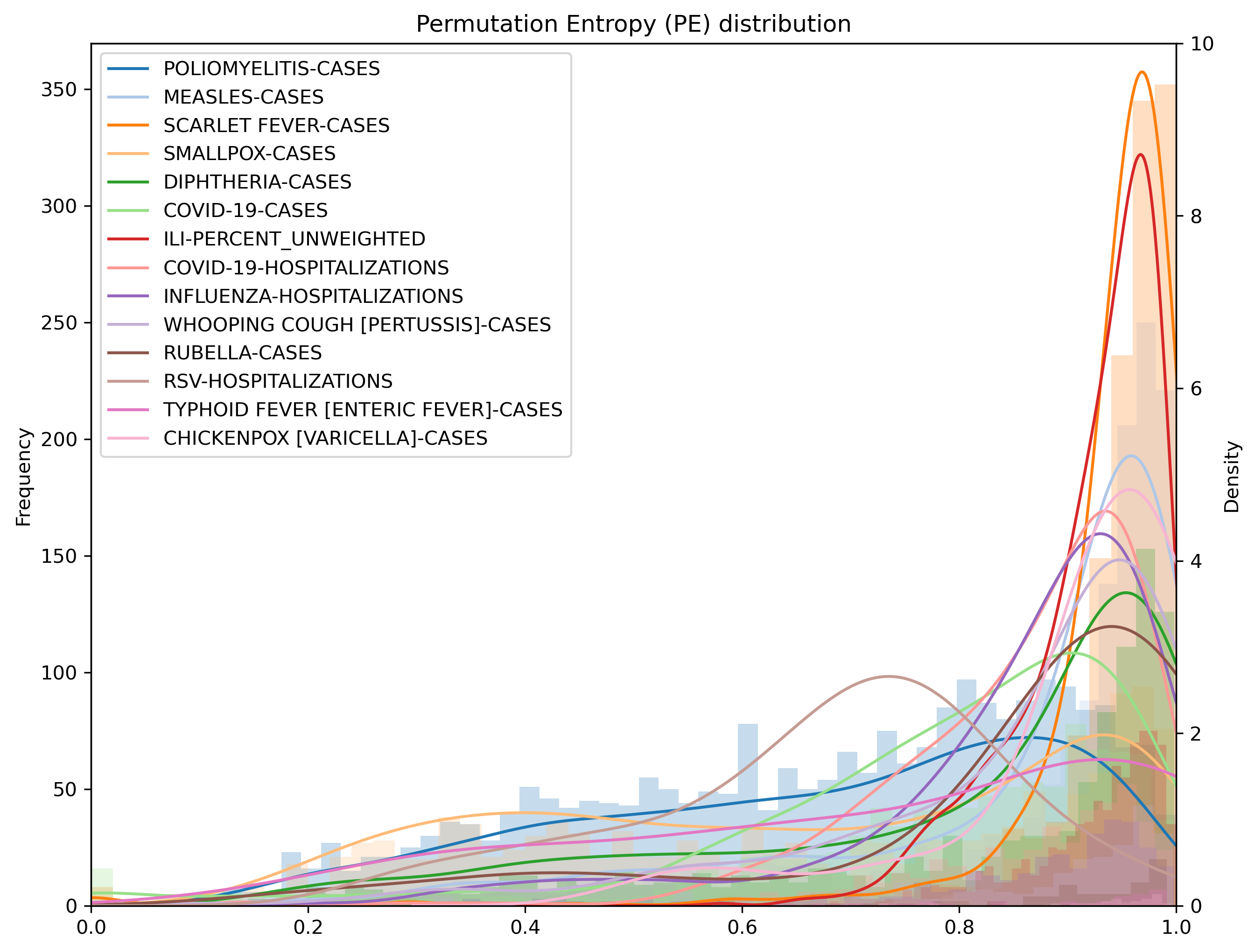}
    \caption{Permutation Entropy distribution}
    \label{fig:permutation_entropy}    
\end{subfigure}
\hfill 
\begin{subfigure}[t]{0.35\textwidth}
\centering
    \includegraphics[width=\textwidth]{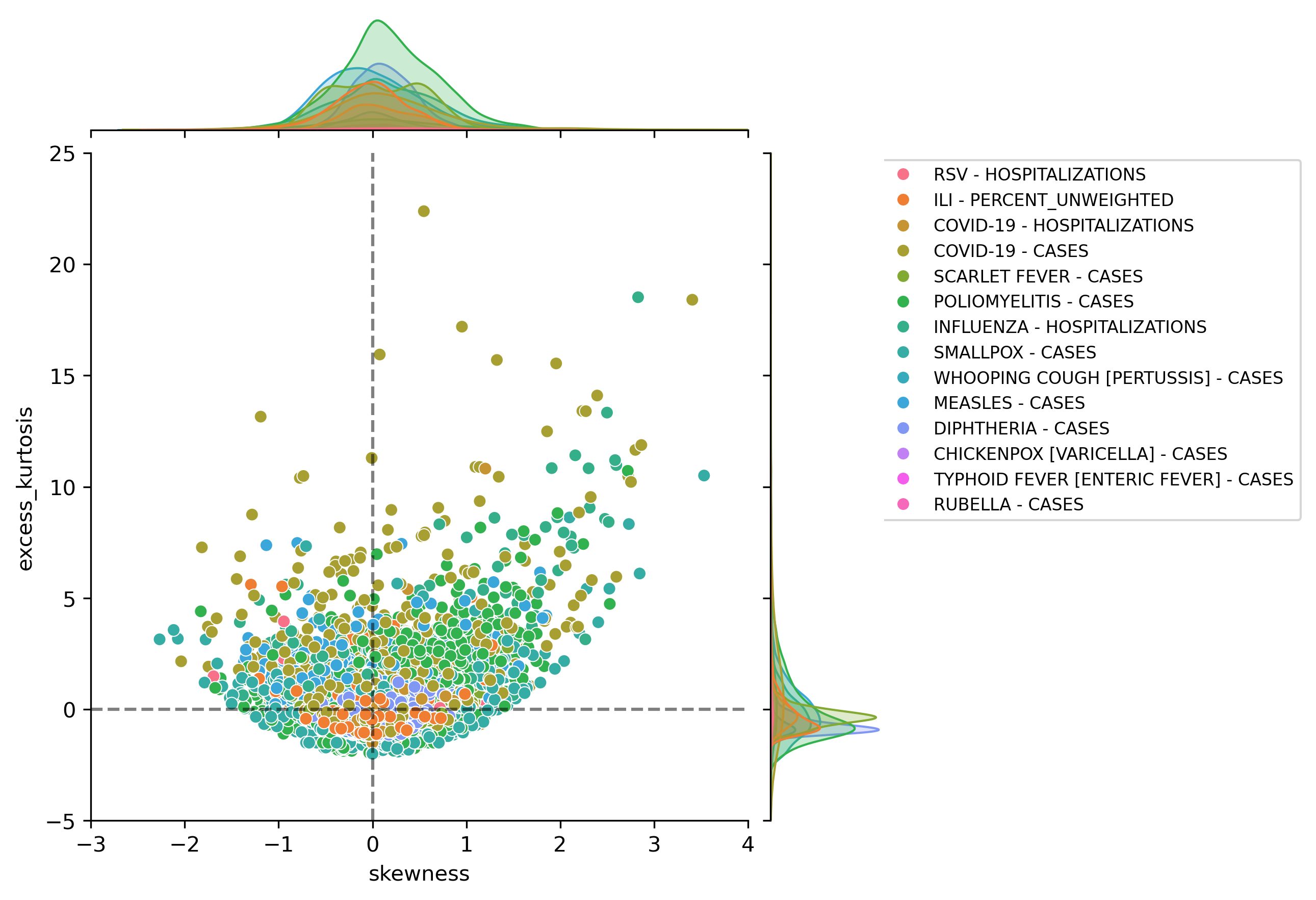}
    \caption{Skewness and Excess Kurtosis}
    \label{fig:kurt_skew}    
\end{subfigure}
    \caption{Analytical measures on the IDOBE dataset}
\end{figure*}

\subsection{Dataset characteristics}
Before describing the insights obtained from the analytical measures, we note that as seen in Table~\ref{tab:stats}, the number of outbreaks vary widely across diseases, depending on the historical trends as well as quality of data capture. For instance, historical vaccine-preventable disease such as poliomyelitis and measles contribute the largest number of outbreaks. Even though spanning fewer years of data capture, due to the global nature of COVID-19 pandemic and seasonal patterns in influenza, they contribute sizeable number of outbreaks as well.  

\subsection{Analytical measures}
Figure~\ref{fig:entropy} shows the distribution of entropies across outbreaks obtained per (disease, outcome) tuple. We observe an universal mode centered around 5, with heterogeneity across diseases. Diseases such as poliomyelitis and smaller have broader entropy distributions, indicating the presence of both sharp (i.e., low entropy) and flatter outbreaks. From Figure~\ref{fig:permutation_entropy}, we note that most outbreaks have high entropy among ordinal patterns of order 3, thus signaling significant noisiness. Certain outbreaks such as those of RSV hospitalizations seem to have lower PE thus hinting at better predictability~\cite{scarpino2019predictability}. Finally, as seen in Figure~\ref{fig:kurt_skew}, most of the outbreaks have negative excess kurtosis (i.e., platykurtic) and positive skew compared to the normal curve. While negative kurtosis confirms the presence of flatter outbreaks, positive skewness indicates the typical ``left-skewed'' nature of epidemic curves with steep inclines before peak and slower declines post-peak. In addition to limited training data, such a characteristic could also result in lower predictability for the early phases for a novel outbreak.

\subsection{Baseline performance}

Table~\ref{tab:forecast_overall_horizon_vertical} shows the performance of the 11 baseline models across different forecast horizons (1- to 4-week ahead) as well as combined performance. We note that performance degrades quickly across horizons. For shorter horizons, both statistical (ETS) and transformer based (TFT) methods perform best. For longer horizones, MLP-based methods have better performance, with MLP performing best for all error metrics for the 4-week ahead. 

Further, we consider post- and pre-peak performance since epidemiological these might vary in relevance for policymakers~\cite{tabataba2017framework}. We observe that the statistical models (ETS) perform best in pre-peak time points, where as transformer-based methods (TFT) perform better for post-peak time points. We also observe that MLP-based methods performance consistently well across both pre- and post-peak phases. This phase-dependent behavior underscores the importance of phase-specific model training and selection~\cite{Aniruddhabigdata}.

Model performance also vary across diseases and outcomes. As seen in Figure~\ref{fig:dis_mod}, all baseline models seem to have poorer performance for poliomyelitis and smallpox cases, with better performance for ILI and Influenza hospitalizations. Average model performance across forecast horizons is shown in Figure~\ref{fig:mod_hor}, with more gradual degradation seen for LSTM and TFT, where statistical models like ARIMA and ETS show a more rapid decline in performance with forecast horizon. 

\section{Discussion}
In this paper, we presented IDOBE, a novel infectious disease outbreak forecasting benchmark ecosystem, comprising preprocessed outbreaks, baseline models, and evaluation metrics. In addition to characterizing the epidemic outbreaks through various analytical measures, we summarize the performance of baseline models within the epidemiological context. Note that the core premise of this work is the need for and ability to segment epidemic surveillance into single-wave outbreaks for benchmarking purposes. We acknowledge that such an approach inherently neglects multi-wave dynamics that may arise from policy changes, spatial heterogeneity and viral evolution. IDOBE, in its current form, is best seen as a benchmark for single-outbreak short-term forecasting and fills an obvious gap in methods development for effective infectious disease response. 

We also focus on univariate forecasting at the level of single location, and disease outcomes. Spatial coupling and models that leverage relationship among multivariate signals have found success in real-time epidemic forecasting. Further, we have not considered mechanistic constraints (e.g., population sizes, disease models) have not been exploited in the current baseline models. We intend to expand IDOBE in future versions to include more sophisticated baseline methods as well as additional epidemiological relevant forecast targets (e.g., peak intensity, total duration). Finally, we also envision incorporating epidemic simulators to produce plausible synthetic outbreaks to augment such benchmark datasets. 

\section{Data and model availability}
IDOBE is openly available at \url{https://github.com/NSSAC/IDOBE}. The repository includes of $(i)$ outbreak data comprising over 10,000 outbreak timeseries across multiple diseases, $(ii)$ tools for data preprocessing, ($iii$) scripts for extracting analytical measures to analyze outbreaks, ($iv$) a suite of trained baseline forecasting models, $(v)$ probabilistic forecast formatting and evaluation scripts.

\section*{In Memoriam}
We would like to acknowledge and mourn the passing of Wilbert Van Panhuis, MD, PhD 1978-2026 whose development of Project Tycho made this work possible.   His clear scientific vision, energy, enthusiasm, and passion for collaborative research was an inspiration for this work, and we are honored to build on his legacy. 

\begin{figure}
    \centering
    \includegraphics[width=0.4\textwidth]{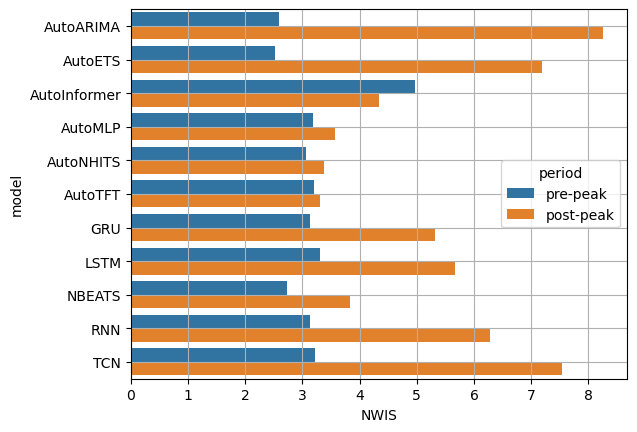}
    \caption{Forecast performance by NWIS pre- and post-outbreak peak date.}
    \label{fig:pre_post_perf}
\end{figure}
\begin{table*}[]
\centering
\caption{Summary of forecast performance across models and horizons. Best (lowest) value per row is boldfaced.}
\label{tab:forecast_overall_horizon_vertical}
\renewcommand{\arraystretch}{1.5}
\resizebox{\linewidth}{!}{%
\begin{tabular}{|c|c|cc|ccc|cc|cccc|}
\hline
\textbf{Horizon} & \textbf{Metric}
& \multicolumn{2}{c|}{\textbf{Statistical}}
& \multicolumn{3}{c|}{\textbf{MLP}}
& \multicolumn{2}{c|}{\textbf{Transformer}}
& \multicolumn{4}{c|}{\textbf{RNN}} \\ \cline{3-13}

& 
& ARIMA & ETS
& MLP & NBEATS & NHITS
& Informer & TFT
& GRU & LSTM & RNN & TCN \\ \hline

\multirow{3}{*}{4-week}
& NWIS
& 6.4 & 5.7
& 3.4 & 3.5 & \textbf{3.3}
& 4.5 & 3.3
& 4.6 & 4.9 & 5.3 & 6.1 \\

& NMSE
& 1.6 & \textbf{1.3}
& 1.5 & 1.4 & 1.4
& 2.5 & 1.4
& 1.6 & 1.6 & 1.5 & 1.5 \\

& MAPE
& 113.3 & 76.2
& 67.3 & 65.8 & 62.9
& 89.7 & \textbf{61.3}
& 72.4 & 74.6 & 80.4 & 115.9 \\ \hline

\multirow{3}{*}{1st wk}
& NWIS
& 4.0 & 4.1
& 3.4 & 2.7 & 2.8
& 4.5 & \textbf{2.7}
& 4.3 & 4.5 & 4.8 & 4.8 \\

& NMSE
& 1.2 & \textbf{1.1}
& 1.6 & 1.3 & 1.4
& 2.6 & 1.2
& 1.5 & 1.6 & 1.5 & 1.4 \\

& MAPE
& 67.1 & 56.1
& 65.3 & 51.5 & 53.7
& 87.3 & \textbf{49.5}
& 68.1 & 68.2 & 76.7 & 89.7 \\ \hline

\multirow{3}{*}{2nd wk}
& NWIS
& 5.6 & 5.2
& 3.4 & 3.2 & \textbf{3.1}
& 4.5 & 3.1
& 4.5 & 4.7 & 5.0 & 5.8 \\

& NMSE
& 1.4 & \textbf{1.3}
& 1.5 & 1.3 & 1.3
& 2.5 & 1.4
& 1.6 & 1.6 & 1.5 & 1.5 \\

& MAPE
& 95.3 & 69.0
& 68.1 & 60.9 & 58.9
& 88.5 & \textbf{58.6}
& 70.7 & 72.2 & 78.0 & 110.0 \\ \hline

\multirow{3}{*}{3rd wk}
& NWIS
& 7.3 & 6.2
& 3.5 & 3.7 & \textbf{3.4}
& 4.6 & 3.5
& 4.7 & 5.0 & 5.4 & 6.8 \\

& NMSE
& 1.8 & 1.5
& 1.4 & \textbf{1.4}
& 1.5
& 2.5 & 1.5
& 1.6 & 1.6 & 1.5 & 1.6 \\

& MAPE
& 130.4 & 82.6
& 67.7 & 70.6 & 66.6
& 91.2 & \textbf{65.5}
& 74.0 & 76.6 & 81.1 & 122.4 \\ \hline

\multirow{3}{*}{4th wk}
& NWIS
& 9.1 & 7.4
& \textbf{3.5} & 4.3 & 3.8
& 4.7 & 3.9
& 5.0 & 5.4 & 5.8 & 7.4 \\

& NMSE
& 2.1 & 1.6
& \textbf{1.3} & 1.4 & 1.6
& 2.5 & 1.5
& 1.5 & 1.6 & 1.5 & 1.6 \\

& MAPE
& 166.2 & 99.9
& \textbf{68.1} & 81.9 & 73.5
& 92.1 & 72.8
& 77.3 & 82.1 & 86.4 & 144.9 \\ \hline

\end{tabular}
}
\end{table*}

\bibliographystyle{plain} 
\bibliography{refs}



\appendix
\counterwithin{figure}{section}

\section{Disease- and Outcome-Specific Baseline Performance}
\begin{figure*}[h!]
    \centering
    \includegraphics[width=\textwidth]{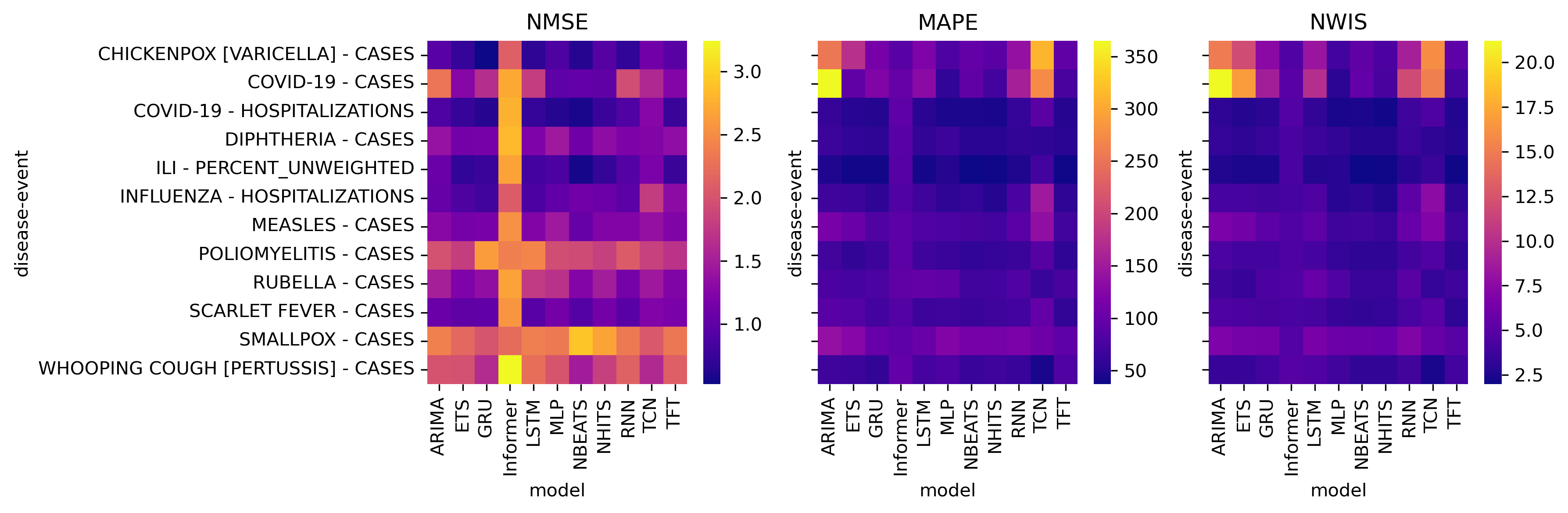}
    \caption{Disease-specific performance across models by different error metrics}
    \label{fig:dis_mod}
\end{figure*}

\begin{figure*}[h!]
    \centering
    \includegraphics[width=\textwidth]{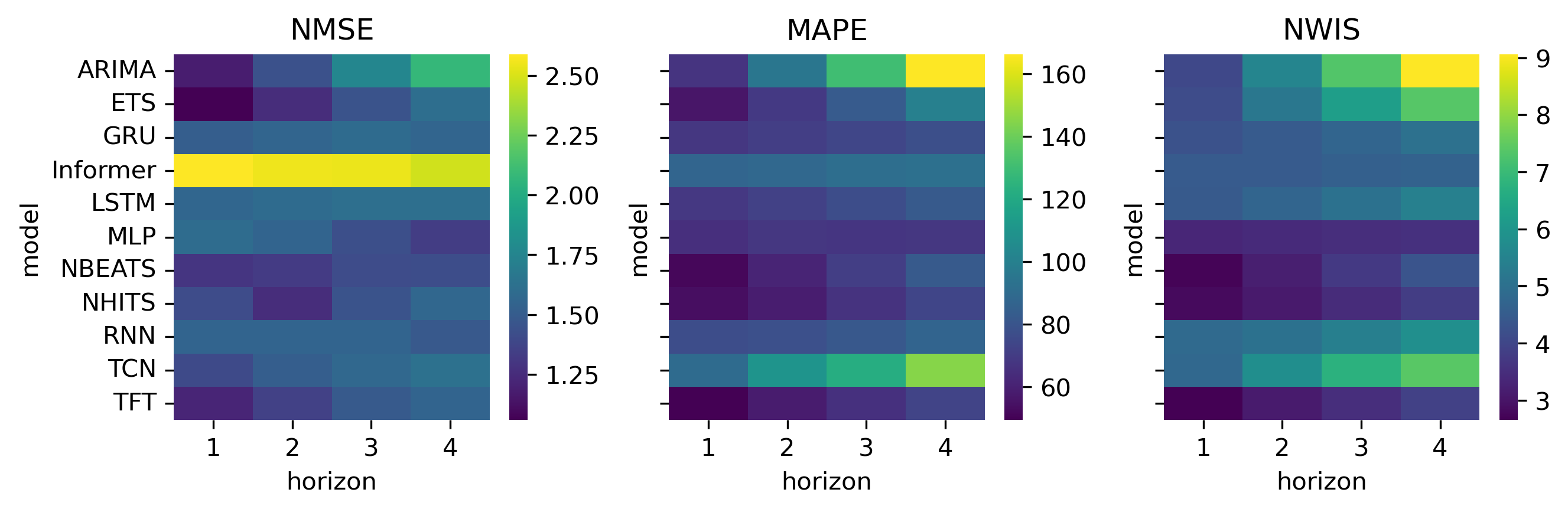}
    \caption{Average model performance across forecast horizons.}
    \label{fig:mod_hor}
\end{figure*}

\end{document}